%% file: VaBe-SamTLPP-TR-2013.tex
\documentclass[letterpaper, 10 pt, conference]{ieeeconf}  

\IEEEoverridecommandlockouts                              
\input{def.tex}
\newcommand{\cut}{\color{blue}}
\newcommand{\add}{\color{blue}}

\title{\LARGE \bf
Sampling-Based Temporal Logic Path Planning
}

\author{Cristian Ioan Vasile and Calin Belta
\thanks{This work was partially supported by the ONR under grants MURI
N00014-09-1051 and MURI N00014-10-10952 and by the NSF under grant NSF
CNS-1035588.}
\thanks{
The authors are with the Division of Systems Engineering, Boston University,
Boston MA, {\tt\small \{cvasile, cbelta\}@bu.edu)}.}%
}

\begin{document}

\maketitle
\thispagestyle{empty}
\pagestyle{empty}

\begin{abstract}

In this paper, we propose a sampling-based motion planning algorithm that finds
an infinite path satisfying a Linear Temporal Logic (LTL) formula over a set of
properties satisfied by some regions in a given environment. The algorithm has
three main features. First, it is incremental, in the sense that the procedure
for finding a satisfying path at each iteration scales only with the number of
new samples generated at that iteration. Second, the underlying graph is sparse,
which guarantees the low complexity of the overall method. Third, it is
probabilistically complete. Examples illustrating the usefulness and the
performance of the method are included.

\end{abstract}

\section{INTRODUCTION}
\label{sec:intro}

Motion planning is a fundamental problem that has received a lot of attention
from the robotics community \cite{Lav06}. The goal is to generate a feasible
path for a robot to move from an initial to a final configuration while avoiding
obstacles. Exact solutions to this problem are intractable, and relaxations
using potential fields, navigation functions, and cell decompositions are
commonly used \cite{Choset.Lynch.ea:05}. These approaches, however, become
prohibitively expensive in high dimensional configuration spaces. Sampling-based
methods were proposed to overcome this limitation.
Examples include the probabilistic roadmap (PRM) algorithm proposed by Kavraki
et.al. in~\cite{Kav96}, which is very useful for multi-query problems, but is
not well suited for the integration of differential constraints. 
In~\cite{KL99}, Kuffner and LaValle proposed rapidly-exploring random trees
(RRT). RRTs grow randomly, are biased to explore ``new'' space~\cite{KL99}
(Voronoi bias), and find solutions quite fast. Moreover, PRM and RRT were shown
to be probabilistically complete~\cite{Kav96,KL99}, but not probabilistically
optimal~\cite{KF-IJRR11}. Karaman and Frazzoli proposed RRT$^*$ and PRM$^*$, the
probabilistically optimal counterparts of RRT and PRM in~\cite{KF-IJRR11}.

Recently, there has been increasing interest in improving the expressivity of
motion planning specifications from the classical  scenario (``Move from A to B
and avoid obstacles.") to richer languages that allow for Boolean and temporal
requirements, e.g., ``Visit A, then B, and then C, in this order infinitely
often. Always avoid D unless E was visited." It has been shown that temporal
logics, such as Linear Temporal Logic (LTL), Computation Tree Logic (CTL),
$\mu$-calculus, and their probabilistic versions (PLTL, PCTL) \cite{Baier08} can
be used as formal and expressive specification languages for robot motion
\cite{kress-gazit:whereswaldo?,Murray2009,bhatia2010sampling,KF-CDC09,DiKlChBe-RAM-2011}.
In the above works, adapted model checking algorithms and automata game
techniques \cite{kress-gazit:whereswaldo?,ChTuBe-ICRA-2012} are used to generate
motion plans and control policies for a finite model of robot motion, which is
usually obtained through an abstraction process based on partitioning the
configuration space \cite{Belta-TRO05}. The main limitation of these approaches
is their high complexity, as both the synthesis and abstraction algorithms scale
at least exponentially with the dimension of the configuration space.

To address this issue, in~\cite{KF-CDC09}, Karaman and Frazzoli proposed a
sampling-based path planning algorithm from specifications given in
deterministic $\mu$-calculus. However, deterministic $\mu$-calculus formulae
have unnatural syntax based on fixed point operators, and are difficult to use
by untrained human operators. In contrast, Linear Temporal Logic (LTL), has very
friendly syntax and semantics, which can be easily translated to natural
language. One idea would be to translate LTL specifications to deterministic
$\mu$-calculus formulae and then proceed with generation of motion plans as
in~\cite{KF-CDC09}.
However, there is no known procedure to transform an LTL formula $\phi$ into a
$\mu$-calculus formula $\Psi$ such that the size of $\Psi$ is polynomial in the
size of $\phi$ (for details see~\cite{CGR10}).

In this paper, we propose a sampling-base path planning algorithm that finds an
infinite path satisfying an LTL formula over a set of properties that hold at
some regions in the configuration space. The procedure is based on the
incremental construction of a transition system followed by the search for one
of its satisfying paths. One important feature of the algorithm is that, at a
given iteration, it only scales with the number of samples and transitions added
to the transitions system at that iteration.
This, together with a notion of ``sparsity" that we define and enforce on the
transition system, play an important role in keeping the overall complexity at a
manageable level. In fact, we show that, under some mild assumptions, our
definition of sparsity leads to the best possible complexity bound for finding a
satisfying path. Finally, while the number of samples increases, the probability
that a satisfying path is found approaches 1, i.e., our algorithm is
probabilistically complete.

Among the above mentioned papers, the closest to this work is~\cite{KF-CDC09}.
As in this paper, the authors of~\cite{KF-CDC09} can guarantee probabilistic
completeness and scalability with added samples only at each iteration of their
algorithm. However, in~\cite{KF-CDC09}, the authors employ the fixed point
(Knaster-Tarski) theorem to find a satisfying path. Their method is based on
maintaining a ``product'' graph between the transition system and every
sub-formula of their deterministic $\mu$-calculus specification and checking for
reachability and the existence of a ``type'' of cycle on the graph. On the other
hand, our algorithm maintains the product automaton between the transition
system and a \buchi automaton corresponding to the given LTL specification. Note
that, as opposed to LTL model checking \cite{Baier08}, we use a modified version
of product automaton that ensures reachability of the final states. Moreover, we
impose that the states of the transition system be bounded away from each other
(by a given function decaying in terms of the size of the transition system).
Sparseness is also explored by Dobson and Berkis in~\cite{Bekris13} for PRM
using different techniques.

{\cut
Our long term goal is to develop a computational framework for automatic
deployment of autonomous vehicles from rich, high level specifications that
combine static, a priori known information with dynamic, locally sensed events.
An example is search and rescue in a disaster relief scenario: an unmanned
aircraft is required to keep on photographing some known affected regions and
uploading the photos at a known base region. While executing this (global)
mission, the aircraft uses its sensors to (locally) identify survivors and
fires, with the goal of immediately providing medical assistance to the
survivors and extinguishing the fires. The main challenge in this problem is to
generate control strategies that guarantee the satisfaction of the global
specification in long term while at the same time correctly reacting to locally
sensed events. The algorithm proposed in this paper solves the first part of
this problem, i.e., the generation of a motion plan satisfying the global
specification.
}


\section{PRELIMINARIES}
For a finite set $\Sigma$, we use $\card{\Sigma}$ and $\spow{\Sigma}$ to
denote its cardinality and power set, respectively. {\cut $\emptyset$ denotes
the empty set.}

\begin{definition}[Deterministic Transition System]
A deterministic transition system (DTS) is a tuple
$\TS = (X, x_0, \Delta, \Pi, h)$, where:
\begin{itemize}
   \item $X$ is a finite set of states;
   \item $x_0 \in X$ is the initial state;
   \item $\Delta \subseteq X \times X$ is a set of transitions;
   \item $\Pi$ is a set of properties (atomic propositions);
   \item $h : X \ra \spow{\Pi}$ is a labeling function.
\end{itemize}
\end{definition}

We denote a transition $(x, x') \in \Delta$ by $x \ra_\TS x'$. A
\emph{trajectory} (or run) of the system is an infinite sequence of states
$\BF{x} = x_0 x_1 \ldots$ such that $x_k \ra_\TS x_{k+1}$ for all $k \geq 0$. A
state trajectory $\BF{x}$ generates an \emph{output trajectory} $\BF{o} = o_0
o_1 \ldots$, where $o_k = h(x_k)$ for all $k \geq 0$.
{\cut The absence of inputs (control actions) in a DTS implicitly means that a
transition $(x, x') \in \Delta$ can be chosen deterministically at every state $x$.
}

\begin{figure}[!htbp]
	\centering
		\includegraphics[scale=0.35]{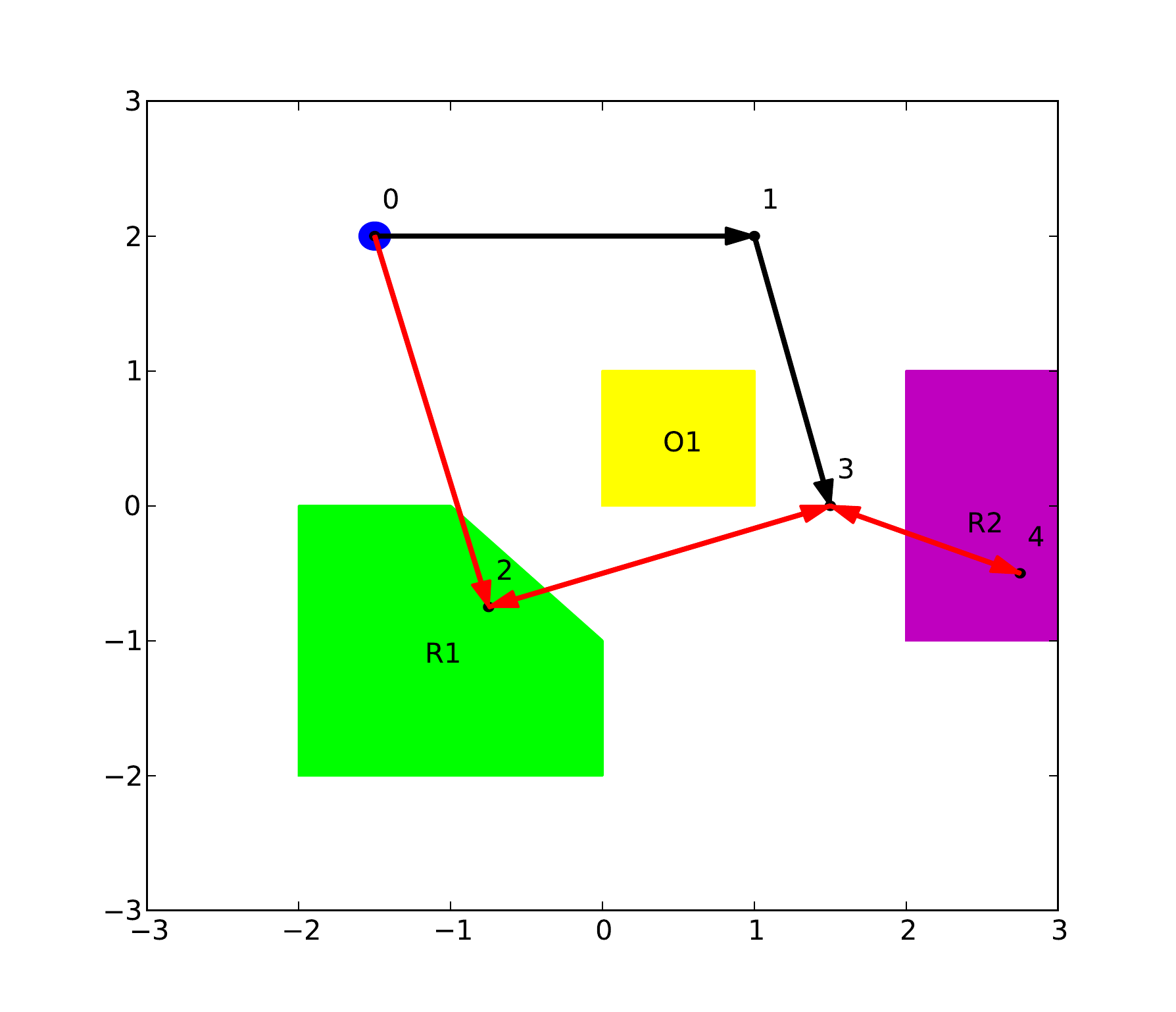}
	\caption{{\cut A simple map with three features: an obstacle $O_1$ and two
	regions of interest $R_1$ and $R_2$. The mission specification is $\phi = \Always ( \Event
	(R_1 \andltl \Event R_2) \andltl \notltl O_1)$. The initial position of the
	robot is marked by the blue disk. The graph (in black and red) represents the
	generated transition system $\TS$. The red arrows specify a satisfying
	trajectory composed of a prefix $[0, 2, 3]$ and infinitely many repetitions of
	the suffix $[4, 3, 2, 3]$.}
	}
	\label{fig:case-simple}
\end{figure}

A Linear Temporal Logic (LTL) formula over a set of properties (atomic
propositions) is defined using standard Boolean operators, $\notltl$ (negation),
$\andltl$ (conjunction) and $\orltl$ (disjunction), and temporal operators,
$\Next$ (next), $\Until$ (until), $\Event$ (eventually), $\Always$ (always). The
semantics of LTL formulae over $\Pi$ are given with respect to infinite words
over $2^\Pi$, such as the output trajectories of the DTS defined above. Any
infinite word satisfying a LTL formula can be written in the form of a finite
prefix followed by infinitely many repetitions of a suffix.
{\cut Verifying whether all output trajectories of a DTS with set of
propositions $\Pi$ satisfy an LTL formula over $\Pi$ is called LTL model checking.
}
LTL formulae can be used to describe rich mission specifications. For example,
formula $\Always ( \Event (R_1 \andltl \Event R_2) \andltl \notltl O_1)$
specifies a persistent surveillance task: ``visit regions $R_1$ and $R_2$
infinitely many times and always avoid obstacle $O_1$''  (see
Figure~\ref{fig:case-simple}). Formal definitions for the LTL syntax, semantics,
and model checking can be found in ~\cite{Baier08}.

\begin{definition}[\buchi Automaton]
A (nondeterministic) \buchi automaton is a tuple $\BA = (S_\BA, S_{\BA_0},
 \Sigma, \delta, F_\BA)$, where:
\begin{itemize}
    \item $S_\BA$ is a finite set of states;
    \item $S_{\BA_0} \subseteq S_\BA$ is the set of initial states;
    \item $\Sigma$ is the input alphabet;
    \item $\delta : S_\BA \times \Sigma \ra \spow{S_\BA}$ is the transition
    function;
    \item $F_\BA \subseteq S_\BA$ is the set of accepting states.
\end{itemize}
\end{definition}

A transition $(s, s') \in \delta(s, \sigma)$ is denoted by $s \ras{\sigma}_\BA
s'$. A trajectory of the \buchi automaton $s_0 s_1 \ldots$ is generated by an
infinite sequence of symbols $\sigma_0 \sigma_1 \ldots$ if $s_0 \in S_{\BA_0}$
and $s_k \ras{\sigma_k} s_{k+1}$ for all $k \geq 0$. A input infinite sequence
over $\Sigma$ is said to be accepted by a \buchi automaton $\BA$ if it generates
at least one trajectory of $\BA$ that intersects the set $F_\BA$ of accepting
states infinitely many times.

It is shown in~\cite{Baier08} that for every LTL formula $\phi$ over $\Pi$ there
exists a \buchi automaton $\BA$ over alphabet $\Sigma = \spow{\Pi}$ such that
$\BA$ accepts all and only those infinite sequences over $\Pi$ that satisfy
$\phi$. There exist efficient algorithms that translate LTL formulae into \buchi
automata~\cite{GaOd01}.

{\cut
Note, that the converse is not true, there are some \buchi automata for which
there is no corresponding LTL formulae. However, there are logics such as
deterministic $\mu$-calculus which are in 1-to-1 correspondence with the set of
languages accepted by \buchi automata.
}

Model checking a DTS against an LTL formula is based on the construction of the
product automaton between the DTS and the \buchi automaton corresponding to the
formula. In this paper, we used a modified definition of the product automaton
that is optimized for incremental search of a satisfying run. Specifically, the
product automaton is defined such that all its states are reachable from the set
of initial states.

\begin{definition}[Product Automaton]
\label{def:pa}
Given a DTS $\TS = (X, x_0, \Delta, \Pi, h)$ and a \buchi automaton
$\BA = (S_\BA, S_{\BA_0}, \spow{\Pi}, \delta_\BA, F_\BA)$, their product
automaton, denoted by $\PA = \TS \times \BA$, is a tuple
$\PA = (S_\PA, S_{\PA_0}, \Delta_\PA, F_\PA)$ where:
\begin{itemize}
    \item $S_{\PA_0} = \{ x_0 \} \times S_{\BA_0}$ is the set of initial states;
    \item $S_\PA \subseteq X \times S_\BA $ is a finite set of states which are
    reachable from some initial state: for every $(x^*, s^*) \in S_\PA$ there
    exists a sequence of $\BF{x} = x_0 x_1 \ldots x_n x^*$, with $x_k \ra_\TS
    x_{k+1}$ for all $0 \leq k < n$ and $x_n \ra_\TS x^*$, and a sequence
    $\BF{s} = s_0 s_1 \ldots s_n s^*$ such that $s_0 \in S_{\BA_0}$, $s_k
    \ras{h(x_k)}_\BA s_{k+1}$ for all $0 \leq k < n$ and $s_n \ras{h(x_n)}_\TS
    s^*$;
    \item $\Delta_\PA \subseteq S_\PA \times S_\PA$ is the set of transitions,
    defined by:
$((x, s), (x', s')) \in \Delta_\PA$ iff $x \ra_\TS x'$ and $s \ras{h(x)}_\BA
s'$;
    \item $F_\PA = (X \times F_\BA) \cap S_\PA$ is the set of accepting
     states.
\end{itemize}
\end{definition}

A transition in $\PA$ is denoted by $(x, s) \ra_\PA (x', s')$ if $((x, s), (x',
s')) \in \Delta_\PA$. A trajectory $\BF{p} = (x_0, s_0) (x_1, s_1) \ldots$ of
$\PA$ is an infinite sequence, where $(x_0, s_0) \in S_{\PA_0}$ and $(x_k, s_k)
\ra_\PA (x_{k+1}, s_{k+1})$ for all $k \geq 0$. Such a trajectory is said to be
accepting if and only if it intersects the set of final states $F_\PA$
infinitely many times. It follows by construction that a trajectory $\BF{p} =
(x_0, s_0) (x_1, s_1) \ldots$ of $\PA$ is accepting if and only if the
trajectory $s_0 s_1 \ldots$ is accepting in $\BA$. As a result, a trajectory of
$\TS$ obtained from an accepting trajectory of $\PA$ satisfies the given
specification encoded by $\BA$.
For $x \in X$, we define $\beta_\PA(x) = \{ s \in S_\BA : (x, s) \in S_\PA \}$
as the set of \buchi states that correspond to $x$ in $\PA$. Also, we denote the
projection of a trajectory $\BF{p} = (x_0, s_0) (x_1, s_1) \ldots$ onto $\TS$ by
$\gamma_\TS(\BF{p}) = x_0 x_1 \ldots$. A similar notation is used for
projections of finite trajectories.

For both DTS and automata, we use $\card{\cdot}$ to denote size, which is the
cardinality of the corresponding set of states. A state of a DTS or an automaton
is called non-blocking if it has at least one outgoing transition.

\section{PROBLEM FORMULATION AND APPROACH}

Let $\CA{D} \subset \BB{R}^n$  be a compact set denoting the configuration space
of a robot.
Let $\CA{R}$ be a set of disjoint regions in $\CA{D}$ and $\Pi$ be a set of
properties of interest corresponding to these regions. A map $\sim : \CA{R} \ra
2^{\Pi}$ specifies how properties are associated to the regions. Throughout this
paper, we will assume that $\CA{R}$ is composed of connected sets with non-empty
interior, which implies they have non-zero Lebesgue measure (i.e. all regions of
interest have full dimension). Also, all connected sets in $\CA{D} \setminus
\bigcup_{R \in \CA{R}} R$ have full dimension.
Examples of properties include ``obstacle", ``target", ``drop-off", etc. (see
Fig.~\ref{fig:case-simple}).
{\cut
In practice, the regions and properties are defined in the workspace, and then
mapped to the configuration space by using standard techniques
\cite{Choset.Lynch.ea:05}. 
}

\begin{problem}
\label{pb:general}
Given an environment described by $(\CA{D}, \CA{R}, \Pi, \sim)$, the initial
configuration of the robot $x_0 \in \CA{D}$ and an LTL formula $\phi$ over the
set of properties $\Pi$, find a satisfying (infinite) path for the robot
originating at $x_0$.
\end{problem}

A possible approach to Problem~\ref{pb:general} is to construct a partition of
the configuration space that contains the regions of interest as elements of the
partition. By using input - output linearizations and vector field assignments
in the regions of the partition, it was shown that ``equivalent" abstractions in
the form of finite (not necessarily deterministic) transition systems can be
constructed for a large variety of robot dynamics that include car-like vehicles
and quadrotors \cite{Belta-TRO05,LinLav09,UlMaOiHuBe-ICRA-2013}. Model checking
and automata game techniques can then be used to control the abstractions from
the temporal logic specification \cite{KB-TAC08-LTLCon}.
The main limitation of this approach is its high complexity, as both the
synthesis and abstraction algorithms scale at least exponentially with the
dimension of the configuration space.

In this paper, we propose a sampling-based approach that can be summarized as follows:
(1) the LTL formula $\phi$ is translated to the \buchi automaton $\BA$; (2) a
transition system $\TS$ is incrementally constructed from the initial position
$x_0$ using an RRG-based algorithm; (3) concurrently with (2), the product
automaton $\PA = \TS \times \BA$ is updated and used to check if there is a
trajectory of $\TS$ that satisfies $\phi$. As it will become clear later, our
proposed algorithm is \emph{probabilistically complete}~\cite{Lav06,KF-IJRR11}
(i.e., it finds a solution with probability 1 if one exists and the number of
samples approaches infinity) and the resulting transition system is
\emph{sparse} (i.e., its states are ``far" away from each other).

\section{PROBLEM SOLUTION}
\label{sec:solution}

The starting point for our solution to Problem~\ref{pb:general} is the RRG
algorithm, which is an extension of RRT~\cite{KF-IJRR11} that maintains a
digraph instead of a tree, and can therefore be used as a model for general
$\omega$-regular languages~\cite{KF-CDC09}. However, we modify the RRG to obtain
a ``sparse'' transition system that satisfies a given LTL formula. More
precisely, a transition system $\TS$ is ``sparse'' if the minimum distance
between any two states of $\TS$ is greater than a prescribed function dependent
only on the size of $\TS$ ($\min_{x, x' \in \TS} \normeucl{x-x'} \geq
\eta(\card{\TS})$). The distance used to define sparsity is inherited from the
underlying configuration space and is not related to the graph theoretical
distance between states in $\TS$. Throughout this paper, we will assume that
this distance is Euclidean.

{\cut
As stated in Section~\ref{sec:intro}, sparsity of $\TS$ is desired {\add since}
the solution to Problem~\ref{pb:general} will be the off-line part of a more
general  procedure that will combine global and local temporal logic
specifications. 
Sparseness also plays an important role in establishing the complexity bounds
for the incremental search algorithm (see Section~\ref{sec:inc-mc}).
}

\subsection{Sparse RRG}

We first briefly introduce the functions used by the algorithm. 

\paragraph{Sampling function} The algorithm has access to a sampling function
$Sample : \BB{N} \ra \CA{D}$, which generates independent and identically
distributed samples from a given distribution $P$. We assume that the support of
$P$ is the entire configuration space $\CA{D}$.

\paragraph{Steer function} The steer function $Steer : \CA{D} \times \CA{D} \ra
\CA{D}$ is defined based on the robot's dynamics.~\footnote{In this paper, we
will assume that we have access to such a function. For more details about
planning under differential constraints see~\cite{Lav06}.} Given a
configurations $x$ and goal configuration $x_g$, it returns a new configuration
$x_n$ that can be reached from $x$ by following the dynamics of the robot and
that satisfies $\normeucl{x_n - x_g} < \normeucl{x - x_g}$.

\paragraph{Near function} $Near : \CA{D} \times \BB{R} \ra \spow{X}$ is a
function of a configuration $x$ and a parameter $\eta$, which returns the set of
states from the transition system $\TS$ that are at most at $\eta$ distance away
from $x$. In other words, $Near$ returns all states in $\TS$ that are inside the
n-dimensional sphere of center $x$ and radius $\eta$.

\begin{figure}[!htbp]
	\centering
		\includegraphics[scale=0.35]{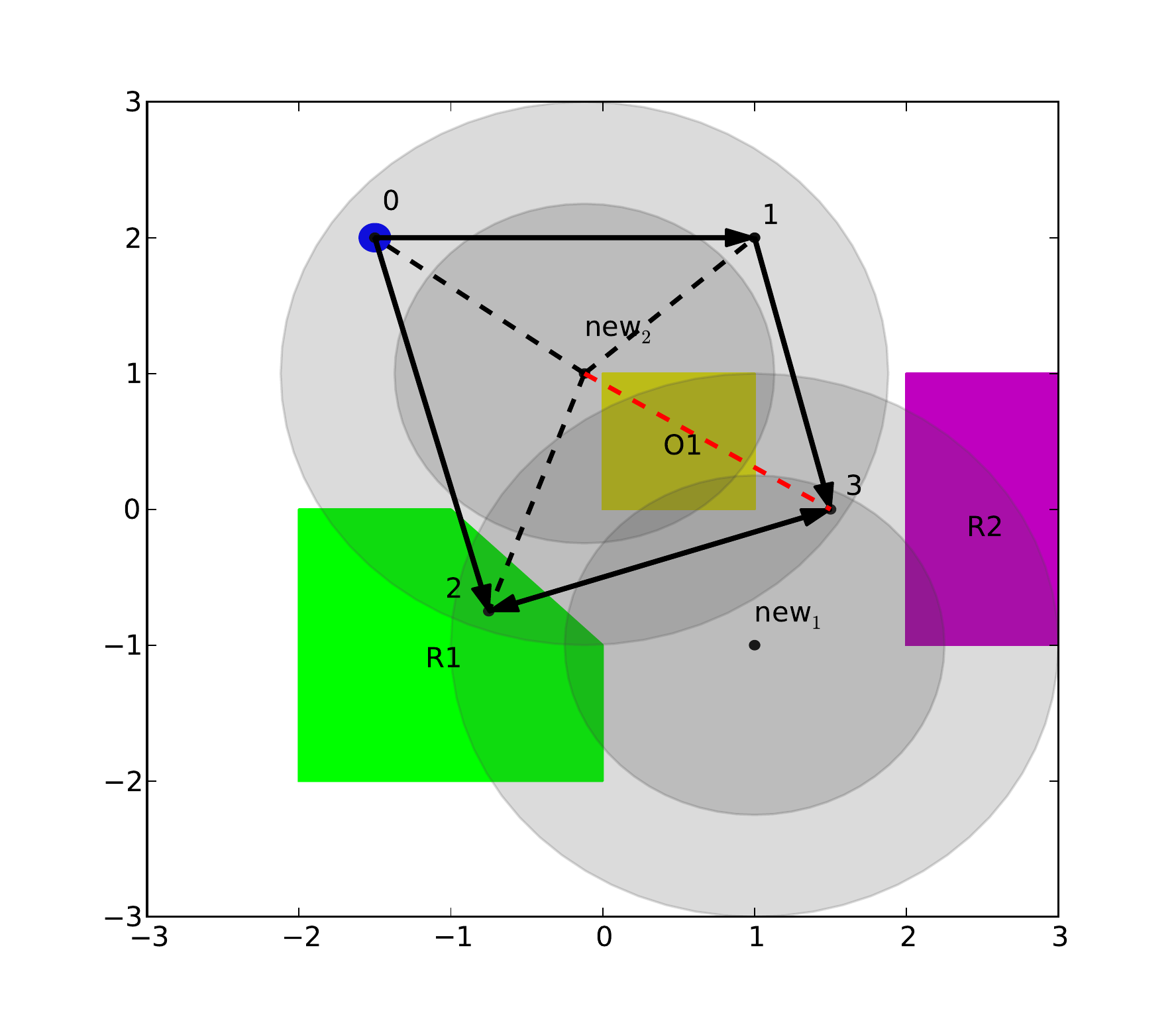}
	\caption{A simple map with three features: an obstacle $O_1$ and two regions
	$R_1$, $R_2$. The robot is assumed to be a fully actuated point. At the current
	iteration the states of $\TS$ are $\{0, 1, 2, 3\}$. The transitions of $\TS$
	are represented by the black arrows. The initial configuration is $0$ and is
	marked by the blue disk. The radii of the dark gray (inner) disks and the light
	gray (outer) disks are $\eta_1$ and $\eta_2$, respectively.
A new sample $new_1 \in \CA{D}$ is generated, but it will not be considered as a
potential new state of $\TS$, because it is within $\eta_1$ distance from state 
3 ($Far(new_1,\eta_1,\eta_2) = \emptyset$). Another sample $new_2 \in \CA{D}$ is
generated, which is at least $\eta_1$ distance away from all states in $\TS$. In
this case, $Far(new_2,\eta_1,\eta_2) = \{0, 1, 2, 3\}$ and the algorithm
attempts to create transition to and from the new sample $new_2$. The
transitions $\{(new_2, 0), (0, new_2), (new_2, 1), (1, new_2), (new_2, 2),
(2, new_2) \}$ (marked by black dashed lines) are added to $\TS$, because all
these transitions correspond to simple line segments ($isSimpleSegment$ returns 
1 for all of them).
{\cut For example, $isSimpleSegment(new_2, 0)=1$, because $new_2$ and $0$ belong
to the same region (the free space region) and $[new_2, 0]$ does not intersect any
other region. $isSimpleSegment(new_2, 2)=1$, because $[new_2, 2]$ crosses the
boundary between the free space region and region $R_1$ once.
}
On the other hand, the transitions  $\{(new_2, 3), (3, new_2)\}$ (marked by red
dashed lines) are not added to $\TS$, since they pass over the obstacle $O_1$.
{\cut In this case, $isSimpleSegment(3, new_2)=0$, because $3$ and $new_2$ are
in the same region, but $[3, new_2]$ crosses the boundary of $O_1$ twice.
}
}
	\label{fig:ranges}
\end{figure}

\paragraph{Far function} $Far : \CA{D} \times \BB{R} \times \BB{R} \ra \spow{X}$
is a function of a configuration $x$ and two parameters $\eta_1$ and $\eta_2$.
It returns the set of states from the transition system $\TS$ that are at most
at $\eta_2$ distance away from $x$. However, the difference from the $Near$
function is that $Far$ returns an empty set if any state of $\TS$ is closer to
$x$ than $\eta_1$.
{\cut
Geometrically, this means that $Far$ returns a non-empty set for a given state
$x$ if there are states in $\TS$ which are inside the $n$-dimensional sphere of
center $x$ and radius $\eta_2$ and all states of $\TS$ are outside the sphere
with the same center, but radius $\eta_1$.}
Thus, $x$ has to be ``far'' away from all states in its immediate neighborhood
(see Figure~\ref{fig:ranges}). This function is used to achieve the
``sparseness'' of the resulting transition system.

\paragraph{isSimpleSegment function} $isSimpleSegment : \CA{D} \times \CA{D} \ra
\{0, 1\}$ is a function that takes two configurations $x_1$, $x_2$ in $\CA{D}$
and returns 1 if the line segment $[x_1, x_2]$ ($\{x \in \BB{R}^n : x = \lambda
x_1 + (1 - \lambda) x_2, \lambda \in [0, 1]\}$) is simple, otherwise it returns
0. A line segment $[x_1, x_2]$ is simple if $[x_1, x_2] \subset \CA{D}$ and the
number of times $[x_1, x_2]$ crosses the boundary of any region $R \in \CA{R}$
is at most one.
Therefore, $isSimpleSegment$ returns 1 if either: (1) $x_1$ and $x_2$ belong to
the same region $R$ and $[x_1, x_2]$ does not cross the boundary of $R$ or (2)
$x_1$ and $x_2$ belong to two regions $R_1$ and $R_2$, respectively, and $[x_1,
x_2]$ crosses the common boundary of $R_1$ and $R_2$ once.
$R$ or at most one of $R_1$ and $R_2$ may be a free space region (a connected
set in $\CA{D} \setminus \bigcup_{R \in \CA{R}}{R}$). See
Figure~\ref{fig:ranges} for examples.
In Algorithm~\ref{code:global-str}, a transition is rejected if it corresponds
to a non-simple line segment (i.e. $isSimpleSegment$ function returns 0).
Under this condition, the satisfaction of the mission specification can be
checked by only looking at the properties corresponding to the states of the
transition system.

\paragraph{Bound functions} $\eta_1 : \BB{Z}_+ \ra \BB{R}$ (lower bound) and
$\eta_2 : \BB{Z}_+ \ra \BB{R}$ (upper bound) are functions that define the
bounds on the distance between a configuration in $\CA{D}$ and the states of the
transition system $\TS$ in terms of the size of $\TS$. These are used as
parameters for functions $Far$ and $Near$. We impose $\eta_1(k) < \eta_2(k)$ for
all $k \geq 1$. We also assume that $c \,\eta_1(k) >  \eta_2(k)$, for some
finite $c>1$ and all $k \geq 0$. Also, $\eta_1$ tends to 0 as $k$ tends to
infinity. The rate of decay of $\eta_1(\cdot)$ has to be fast enough such that a
new sample may be generated. Specifically, the set of all configurations where
the center of an $n$-sphere of radius $\eta_1/2$ may be placed such that it does
not intersect any of the $n$-spheres corresponding to the states in $\TS$ has to
have non-zero measure with respect to the probability measure $P$ used by the
sampling function.
{\cut One conservative upper bound is $\eta_1(k) < \frac{1}{\sqrt{\pi}}
\sqrt[n]{\frac{\mu(\CA{D}) \Gamma(n/2+1)}{k}}$ for all $k \geq 1$, where
$\mu(\CA{D})$ is the total measure (volume) of the configuration space, $n$ is
the dimension of $\CA{D}$, and $\Gamma$ is the gamma function.
This bound corresponds to the case when there is enough space to insert an
$n$-sphere of radius $\eta_1/2$ between every two distinct states of $\TS$.
}
To simplify the notation, we drop the parameter for these functions and assume
that $k$ is always given by the current size of the transition system,
$k=\card{\TS}$.

The goal of the modified RRG algorithm (see Algorithm~\ref{code:global-str}) is
to find a satisfying run, but such that the resulting transition system is
``sparse'', i.e. states are ``sufficiently'' apart from each other. The
algorithm iterates until a satisfying run originating in $x_0$ is found.

At each iteration, a new sample $x_r$ is generated (line 6 in
Algorithm~\ref{code:global-str}). For each state $x$ in $\TS$ which is ``far''
from the sample $x_r$ ($x \in Far(x_r, \eta_1, \eta_2)$), a new configuration
$x'_r$ is computed such that the robot can be steered from $x$ to $x'_r$ and the
distance to $x_r$ is decreased (line 10). The two loops of the algorithm (lines
7--13 and 16--21) are executed if and only if the $Far$ function returns a
non-empty set. However, $x'_r$ is regarded as a potential new state of $\TS$,
and not $x_r$. Thus, the $Steer$ function plays an important role in the
``sparsity'' of the final transition system.
Next, it is checked if the potential new transition $(x, x'_r)$ is a simple
segment (line 9). It is also verified if $x'_r$ may lead to a solution, which is
equivalent to testing if $x'_r$ induces at least one non-blocking state in $\PA$
(see Algorithm~\ref{code:global-inc-mc}). If configuration $x'_r$ and the
corresponding transition $(x, x'_r)$ pass all tests, then they are added to the
list of new states and list of new transitions of $\TS$, respectively (lines
12--13).

After all ``far'' neighbors of $x_r$ are processed, the transition system is
updated. Note that at this point $\TS$ was only extended with states that
explore ``new space''. However, in order to model $\omega$-regular languages the
algorithm must also close cycles. Therefore, the same procedure as before (lines
7--14) is also applied to the newly added states $X'$ (lines 15--21 of
Algorithm~\ref{code:global-str}). The difference is that it is checked if states
from $X'$ can steer the robot back to states in $\TS$ in order to close cycles.
Also, because we know that the states in $X'$ are ``far'' from their neighbors,
the $Near$ function will be used instead of the $Far$ function. The algorithm
returns a (prefix, suffix) pair in $\TS$ obtained by projection from the
corresponding path ($p_0 \ras{*} p_F$) and cycle ($p_F \ras{+} p_F$) in $\PA$,
respectively. The $*$ above the transition symbol means that the length of the
path can be 0, while $+$ denotes that the length of the cycle must be at least
1.

\begin{algorithm}
\footnotesize
\caption{Sparse RRG}
\label{code:global-str}
\DontPrintSemicolon
\KwIn{$\BA$ -- \buchi automaton corresponding to $\phi$}
\KwIn{$x_0$ initial configuration of the robot}
\KwOut{(prefix, suffix) in $\TS$}
\BlankLine

Construct $\TS$ with $x_0$ as initial state\;
Construct $\PA = \TS \times \BA$\;
Initialize $scc(\cdot)$\;

\While{$\neg (x_0 \models \phi)$ $(\equiv \neg (\exists p \in F_\PA$ s.t. $\card{scc(p)} > 1))$}{
  $X' \asgn \emptyset$, $\Delta' \asgn \emptyset$, $\Delta'_\PA \asgn \emptyset$\;
  $x_r \asgn Sample()$\;
  \ForEach{$x \in Far(x_r, \eta_1, \eta_2)$} {
    $x'_r \asgn Steer(x, x_r)$\;
    \If{$isSimpleSegment(x_r, x'_r)$}{
      $added \asgn updatePA(\PA, \BA, (x, x'_r))$\;
      \If{$added$ is True}{
        $X' \asgn X' \cup \{x'_r\}$\;
        $\Delta' \asgn \Delta' \cup \{(x, x'_r)\}$\;
      }
    }
  }
  $\TS \asgn \TS \cup (X', \Delta')$\;
  
  \BlankLine
  $\Delta' \asgn \emptyset$, $\Delta'_\PA \asgn \emptyset$\;
  \ForEach{$x'_r \in X'$}{
    \ForEach{$x \in Near(x'_r, \eta_2)$}{
      \If{$(x = Steer(x'_r, x)) \wedge isSimpleSegment(x'_r, x)$}{
        $added \asgn updatePA(\PA, \BA, (x, x'_r))$\;
        \If{$added$ is True}{
          $\Delta' \asgn \Delta' \cup \{(x'_r, x)\}$\;
        }
      }
    }
  }
  $\TS \asgn \TS \cup (X', \Delta')$\;
}
\Return{$(\gamma_\TS(p_0 \ras{*} p_F), \gamma_\TS(p_F \ras{+} p_F))$, where $p_F \in F_\PA$}
\end{algorithm}

{\cut
In the end, the result is a transition system $\TS$ which captures the general
topology of the environment. In the next section, we will show that $\TS$ also
yields a run that satisfies the given specification.
}

\subsection{Incremental search for a satisfying run}
\label{sec:inc-mc}

The proposed approach of incrementally constructing a transition system raises
the problem of how to efficiently check for a satisfying run at each iteration.
As mentioned in the previous section, the search for satisfying runs is
performed on the product automaton. Note that testing whether there exists a
trajectory of $\TS$ from the initial position $x_0$ that satisfies the given LTL
formula $\phi$ is equivalent to searching for a path from an initial state $p_0$
to a final state $p_F$ in the product automaton $\PA = \TS \times \BA$ and for a
cycle containing $p_F$ of length greater than 1, where $\BA$ is the \buchi
automaton corresponding to $\phi$. If such a path and a cycle are found then
their projection onto $\TS$ represents a satisfying infinite trajectory (line 23
of Algorithm~\ref{code:global-str}). Testing whether $p_F$ belongs to a
non-degenerate cycle (length greater than 1) is equivalent to testing if $p_F$
belong to a non-trivial strongly connected component -- SCC (the size of the SCC
is greater than 1). Checking for a satisfying trajectory in $\PA$ is performed
incrementally as the transition system is modified.

The reachability of the final states from initial ones in $\PA$ is guaranteed by
construction (see Definition~\ref{def:pa}). However, we need to define a
procedure (see Algorithm~\ref{code:global-inc-mc}) to incrementally update $\PA$
when a new transition is added to $\TS$. Consider the (non-incremental) case of
constructing $\PA = \TS \times \BA$. This is done by a traversal of $\bar{\PA} =
(X \times S_\BA, \bar{\Delta}_\PA)$ from all initial states, where $((x, s),
(x', s')) \in \bar{\Delta}_\PA$ if $p \ra p'$ and $s \ras{h(x)} s'$. $\bar{\PA}$
is a product automaton but without the reachability requirement. This suggests
that the way to update $\PA$ when a transition $(x, x')$ is added to $\TS$, is
to do a traversal from all states $p$ of $\PA$ such that $\gamma_\TS(p)=x$.
Also, it is checked if $x'$ induces any non-blocking states in $\PA$ (lines 1-3
of Algorithm~\ref{code:global-inc-mc}). The test is performed by computing the
set $S'_\PA$ of non-blocking states of $\PA$ (line 1) such that $p'\in S'_\PA$
has $\gamma_\TS(p')=x'$ and $p'$ is obtained by a transition from $\{ (x, s) : s
\in\beta_\PA(x) \}$. If $S'_\PA$ is empty then the transition $(x, x')$ of $\TS$
is discarded and the procedure stops (line 3). Otherwise, the product automaton
$\PA$ is updated recursively to add all states that become reachable because of
the states in $S'_\PA$. The recursive procedure is performed from each state in
$S'_\PA$ as follows: if a state $p$ (line 7) is not in $\PA$, then it is added
to $\PA$ together with all its outgoing transitions (line 10) and the recursive
procedure continues from the outgoing states of $p$; if $p$ is in $\PA$ then the
traversal stops, but its outgoing transitions are still added to $\PA$ (line
14). The incremental construction of $\PA$ has the same overall complexity as
constructing $\PA$ from the final $\TS$ and $\BA$, because the recursive
procedure just performes traversals that do not visit states already in $\PA$.
Thus, we focus our complexity analysis on the next step of the incremental
search algorithm.

\begin{algorithm}
\footnotesize
\caption{Incremental Search for a Satisfying Run}
\label{code:global-inc-mc}
\DontPrintSemicolon
\KwIn{$\PA$ -- product automaton}
\KwIn{$\BA$ -- \buchi automaton}
\KwIn{$(x, x')$ -- new transition in $\TS$}
\KwOut{Boolean value -- indicates if $\PA$ was modified}
\BlankLine

$S'_\PA \asgn \{ (x', s') : s \ras{h(x)}_\BA s', s \in \beta_\PA(x), s'$ non-blocking$\}$\;
$\Delta'_\PA \asgn \{((x, s), (x', s')) : s \in \beta_\PA(x), s \ras{h(x)}_\BA s',\ (x', s') \in S'_\PA \}$\;
\If{$S'_\PA \neq \emptyset$}{
    $\PA \asgn \PA \cup (S'_\PA, \Delta'_\PA)$\;
    $stack \asgn S'_\PA$\;
    \While{$stack \neq \emptyset$}{
        $p_1 = (x_1, s_1) \asgn stack.pop()$\;
        \ForEach{$p_2 \in \{ (x_2, s_2) : x_1 \ra_\TS x_2, s_1 \ras{h(x_1)}_\BA s_2 \}$} {
            \If{$p_2 \notin S_\PA$}{
                $\PA \asgn \PA \cup (\{ p_2 \}, \{ (p_1, p_2)\})$\;
                $\Delta'_\PA \asgn \Delta'_\PA \cup \{ (p_1, p_2)\}$\;
                $stack \asgn stack \cup \{ p_2\}$\;
            }
            \ElseIf{$(p_1, p_2) \notin \Delta_\PA$}{
                $\Delta_\PA \asgn \Delta_\PA \cup \{ (p_1, p_2)\}$\;
                $\Delta'_\PA \asgn \Delta'_\PA \cup \{ (p_1, p_2)\}$\;
            }
        }
    }
    updateSCC($\PA$, $scc$, $\Delta'_\PA$)\;
    \Return True
}
\Return False
\end{algorithm}

The second part of the incremental search procedure is concerned with
maintaining the strongly connected components (SCCs) of $\PA$ (line 14 of
Algorithm~\ref{code:global-inc-mc}) as new transitions are added (these are
stored in $\Delta'_\PA$ in Algorithm~\ref{code:global-inc-mc}). To incrementally
maintain the SCCs of the product automaton, we employ the soft-threshold-search
algorithm presented in~\cite{Haeupler2012}. The algorithm maintains a
topological order of the super-vertices corresponding to each SCC. When a new
transition is added to $\PA$, the algorithm proceeds to re-establish a
topological order and merges vertices if new SCCs are formed. The details of the
algorithm are presented in~\cite{Haeupler2012}. The authors also offer insight
about the complexity of the algorithm. They show that, under a mild assumption,
the incremental algorithm has the best possible complexity bound.

Incrementally maintaining $\PA$ and its SCCs yields a quick way to check if a
trajectory of $\TS$ satisfies $\phi$ (line 4 of
Algorithm~\ref{code:global-str}). {\cut The next theorem establishes the overall
complexity of Algorithm 2.}

\begin{theorem}
The execution time of the incremental search algorithm~\ref{code:global-inc-mc}
is $O(m^\frac{3}{2})$, where $m$ is the number of transitions added to $\TS$ in
Algorithm~\ref{code:global-str}.
\end{theorem}
\begin{remark}
First, note that the execution time of the incremental procedure is better by a
polynomial factor than naively running a linear-time SCC algorithm at each step,
since this will have complexity $O(m^2)$. {\cut The algorithm presented
in~\cite{Haeupler2012} improves the previously best known bound by a logarithmic
factor (for sparse graphs). The proof exploits the fact that the ``sparseness''
(metric) property we defined implies a topological sparseness, i.e., $\TS$ is a
sparse graph.}
\end{remark}
{\cut
\begin{proof}
The proof of the theorem is based on the analysis from~\cite{Haeupler2012} of
incremental SCC algorithms. Haeupler et.al. show that any incremental algorithm
that satisfies a ``local'' property must take at least $\Omega(n \sqrt{m})$
time, where $n$ is the number of nodes in the graph and $m$ is the number of
edges. The ``local'' property is a mild assumption that restricts the algorithm
to reorder only vertices that are affected by the addition of an edge. This
implies that the incremental SCC algorithm has the best possible complexity
bound, in asymptotic sense, if and only if the graph is sparse. A graph is
sparse if the the number of edges is asymptotically the same as the number of
nodes, i.e. $m = O(n)$. What we need to show is that the transition system
generated by Algorithm~\ref{code:global-str} is sparse. Note that although we
run the SCC algorithm on the product automaton, the asymptotic execution time is
not affected by analyzing the transition system instead of the product
automaton, because the \buchi automaton is fixed. This follows from
$\card{S_\PA} \leq \card{S_\BA} \cdot \card{X}$ and $\card{\Delta_\PA} \leq
\card{\delta_\BA} \cdot \card{\Delta}$.

Intuitively, the underlying graph of $\TS$ is sparse, because the states were
generated ``far'' from each other. When a new state is added to $\TS$, it will
be connected to other states that are at least $\eta_1$ and at most $\eta_2$
distance away. Also, all states in $\TS$ are at least $\eta_1$ distance away
from each other. This implies that there is a bound on the density of states.
This bound is related to the kissing number~\cite{Kissing}. The kissing number
is the maximum number of non-overlapping spheres that touch another given
sphere. {\add Using this intuition, the problem of estimating the maximum number
of neighbors of a state can be restated as a sphere packing problem. Given a
state $x$, each neighbor can be thought of as a sphere with radius $\eta_1/2$
and center belonging to the volume delimited by two spheres centered at $x$ and
with radii $\eta_1$ and $\eta_2$, respectively. Since, $\eta_1< \eta_2 < c
\eta_1$ for some $c>1$ it follows, that there will be only a finite number of
spheres which can be placed inside the described volume.} Thus there is a finite
bound on the number of neighbors a state can have, which depends only on the
dimension and shape of the configuration space and the volume between two
concentric spheres of radii $\eta_1$ and $\eta_2$, respectively.
\end{proof}
\begin{remarks}
The bound on the number of neighbors can become very large as the dimension of
the configuration space increases.

As we have seen in the proof, under the ``local'' property assumption, the
incremental search algorithm has the best possible complexity bound. Because we
do the search for a satisfying run using a \buchi automaton, through the product
automaton, and not the LTL formula directly, the proposed method is general
enough to be applied in conjunction with logics (such as $\mu$-calculus), which
are as expressive as the languages accepted by \buchi automata. Also, we do not
expect to obtain search algorithms that are asymptotically faster then the
proposed one (for sparse graphs), since this would violate the lower bound
obtained in~\cite{Haeupler2012} (assuming the ``local'' property).
\end{remarks}
}

{\cut \subsection{Probabilistic completeness}}
The presented RRG-based algorithm retains the probabilistic completeness of RRT,
since the constructed transition system is composed of an RRT-like tree and some
transitions which close cycles.

\begin{theorem}
Algorithm~\ref{code:global-str} is probabilistically complete.
\end{theorem}
{\cut
\begin{proof}(Sketch)
First we start by noting that any word in a $\omega$-regular language can be
represented by a finite prefix and a finite suffix, which is repeated
indefinitely~\cite{Baier08}. This is important, since this shows that a
solution, represented by a transition system, is completely characterized by a
finite number of states. Let us denote by $\bar{X}$ the finite set of states
that define a solution. It follows from the way regions are defined that we can
choose a neighborhood around each state in $\bar{X}$ such that the system can be
steered in one step from all points in one neighborhood to all points in the
next neighborhood. Thus, we can use induction to show that~\cite{KF-ACC12}: (1)
there is a non-zero probability that a sample will be generated inside the
neighborhood of the first state in the solution sequence; (2) if there is a
state in $\TS$ that is inside the neighborhood of the $k$-th state from the
solution sequence, then there is a non-zero probability that a sample will be
generated inside the $k+1$-st state's neighborhood. Therefore, as the number of
samples goes to infinity, the probability that the transition system $\TS$ has
nodes belonging to all neighborhoods of states in $\bar{X}$ goes to 1.
{\add To finish the proof, note that we have to show that the algorithm is
always able to generate samples with the desired ``sparseness'' property.
However, recall that the bound functions must converge to 0 (as the number of
states goes to infinity) fast enough such that the set of configurations for
which ``Far'' function returns a non-empty list has non-zero measure with
respect to the sampling distribution. This concludes the proof.}
\end{proof}
}

\section{IMPLEMENTATION AND CASE STUDIES}

We implemented the algorithms presented in this paper in Python2.7. In this
section, we present some examples in configuration spaces of dimensions 2, 10
and 20. In all cases, we assume for simplicity that the $Steer$ function is
trivial, i.e., there are no actuation constraints at any given configuration.
All examples were ran on an iMac system with a 3.4 GHz Intel Core i7 processor
and 16GB of memory.

\begin{figure}[!htb]
    \centering
    \includegraphics[width=0.35\textwidth]{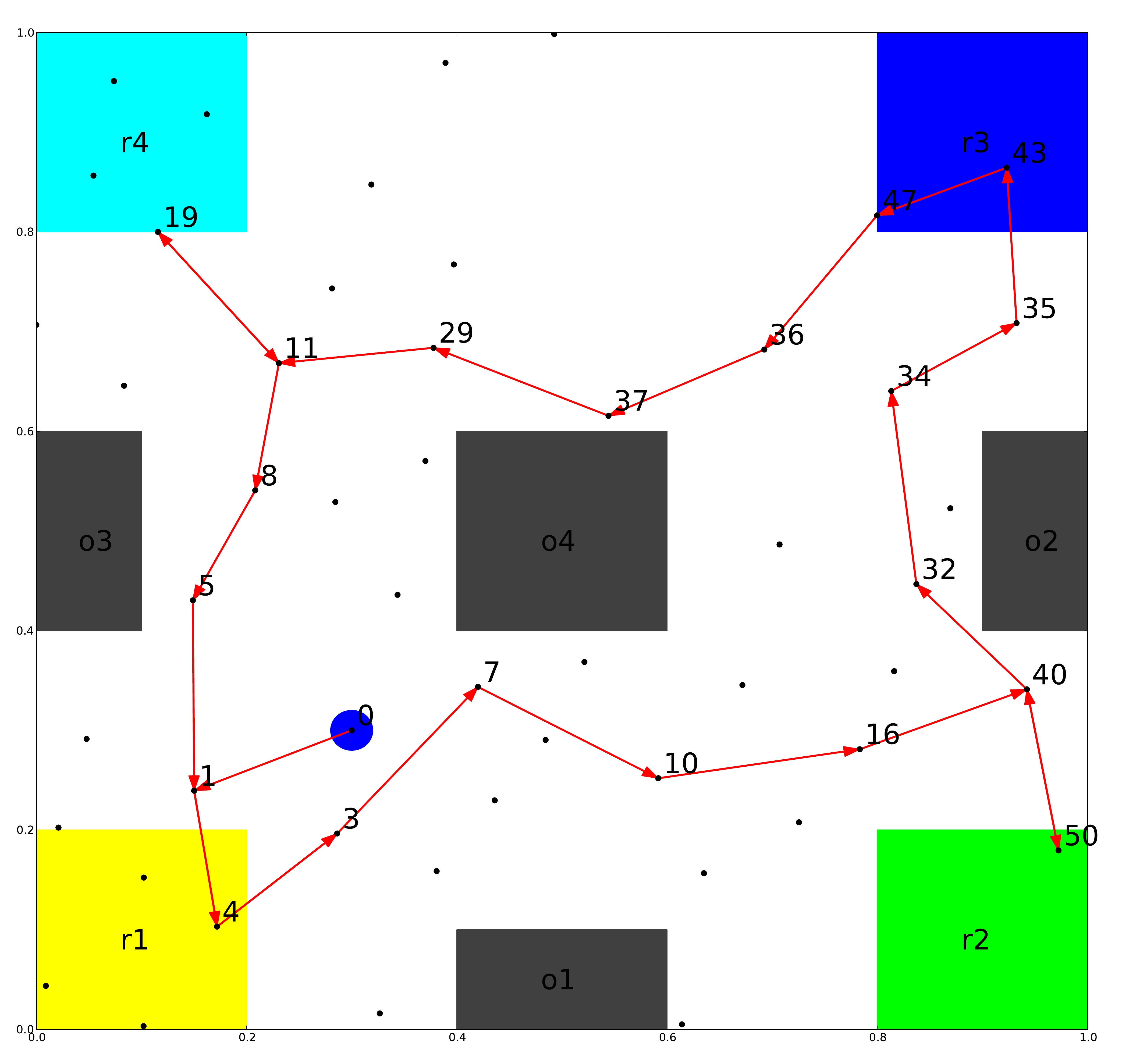}
    \caption{One of the solutions corresponding to Case Study 1: the
specification is to visit all the colored regions labelled $r1$ (yellow), $r2$
(green), $r3$ (blue) and $r4$ (cyan) infinitely often, while avoiding the dark
gray obstacles labelled $o1$, $o2$, $o3$, $o4$. The black dots represent the
states of the transition system $\TS$ (51 states and 264 transitions). The
starting configuration of the robot (the initial state of $\TS$) is denoted by
the blue circle. The red arrows represent the satisfying run (finite prefix,
suffix pair) found by Algorithm~\ref{code:global-str}, which is  composed of 21
states from $\TS$. In this case, the prefix and suffix are [0, 1, 4, 3] and [7,
10, 16, 40, 50, 40, 32, 34, 35, 43, 47, 36, 37, 29, 11, 19, 11, 8, 5, 1, 4, 3],
respectively.}
	\label{fig:case1}
\end{figure}

{\bf Case Study 1:} Consider the configuration space depicted in
Figure~\ref{fig:case1}. The initial configuration is at $(0.3; 0.3)$. The
specification is to visit regions $r1$, $r2$, $r3$ and $r4$ infinitely many
times while avoiding regions $o1$, $o2$, $o3$ and $o4$. The corresponding LTL
formula for the given mission specification is
\begin{eqnarray}\label{eq:phi-1}
\phi_1 &=& \Always( \Event r1 \andltl ( \Event r2 \andltl ( \Event r3 \andltl ( \Event r4 ) ) ) \\
       & & \andltl \notltl (o1 \orltl o2 \orltl o3 \orltl o4)) \notag
\end{eqnarray}

\begin{figure}[!htb]
  \centering
  \subfigure{
    \includegraphics[width=0.2\textwidth]{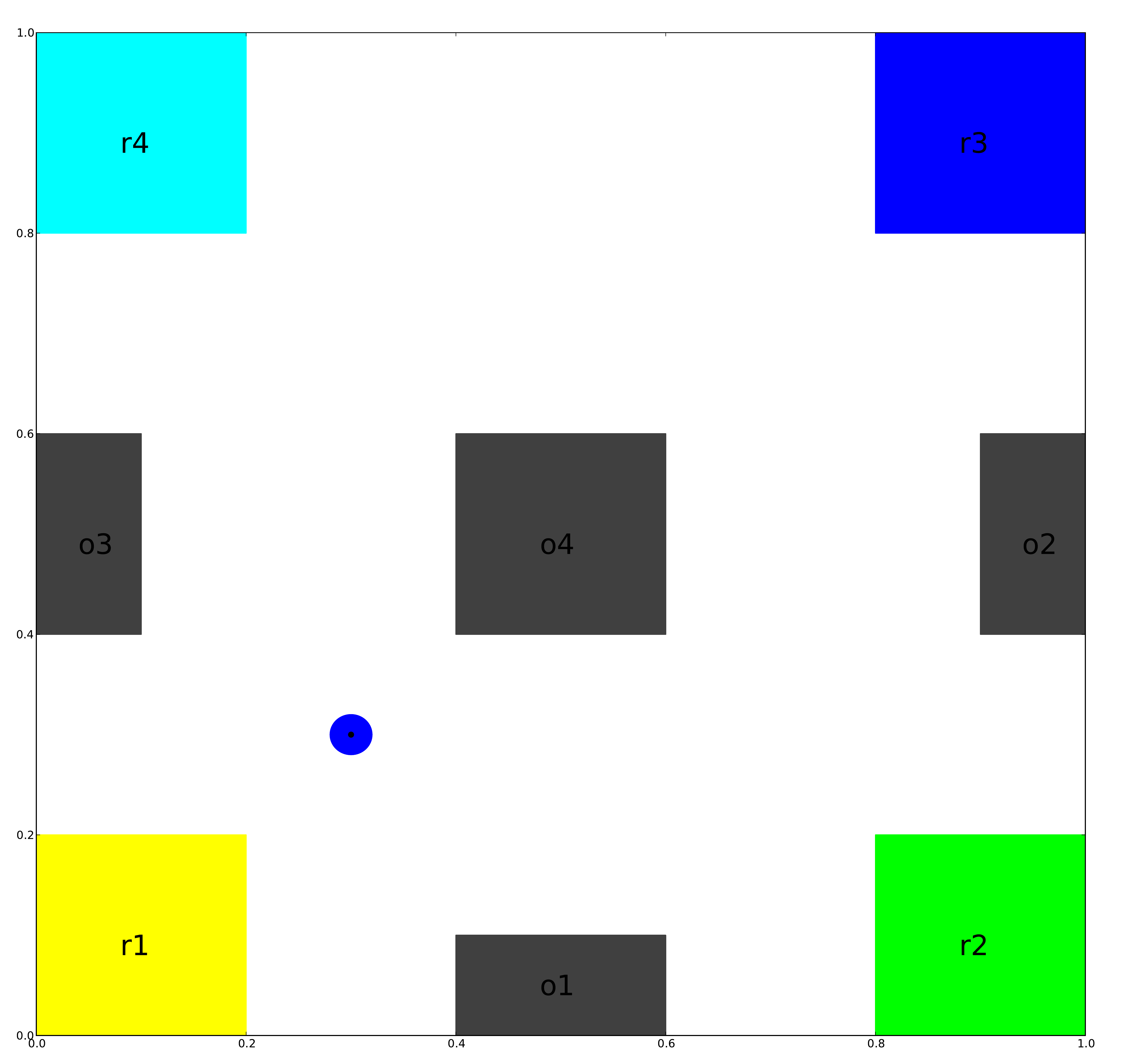}
    \label{fig:case1-s0}
  }
  \subfigure{
    \includegraphics[width=0.2\textwidth]{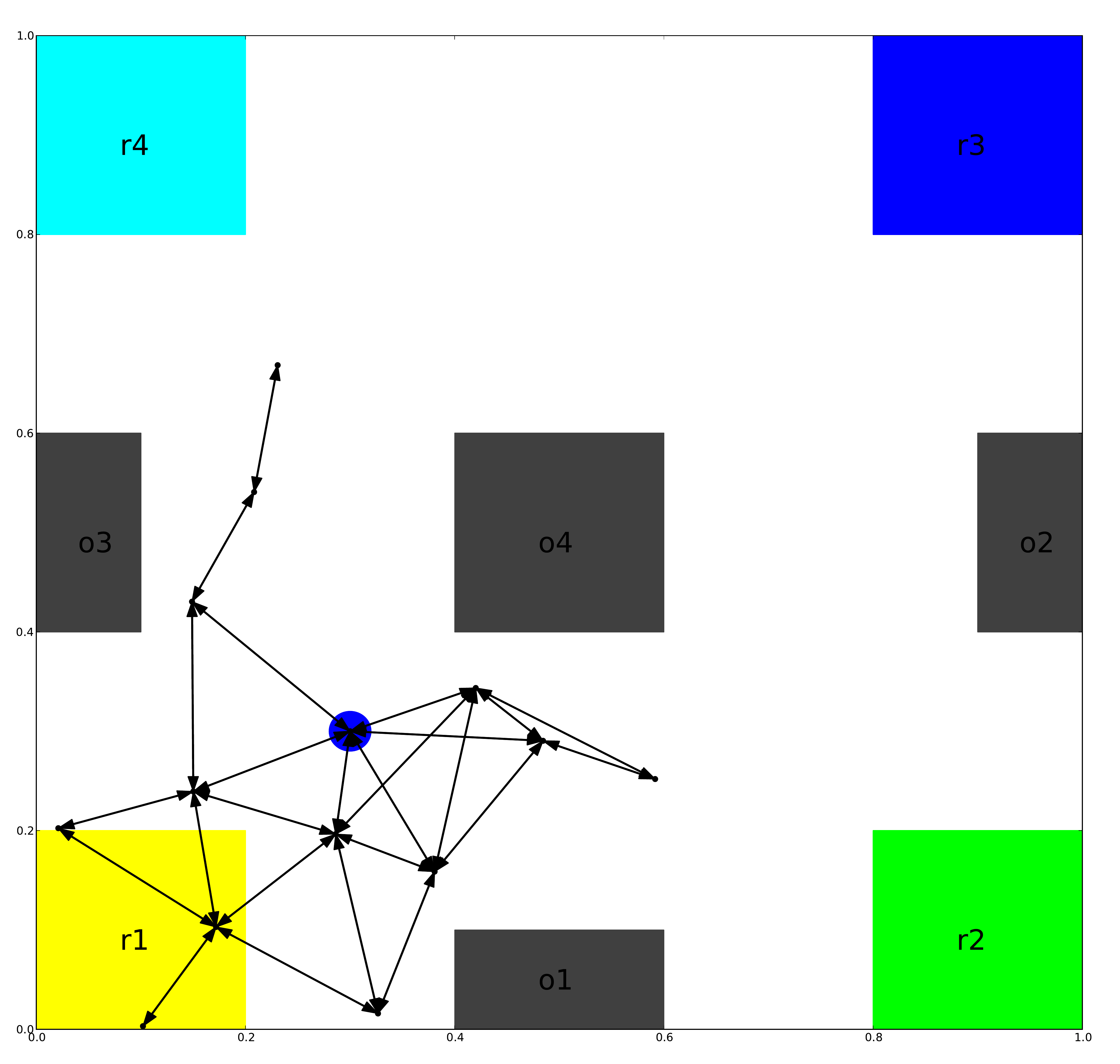}
    \label{fig:case1-s1}
  }
  \subfigure{
    \includegraphics[width=0.2\textwidth]{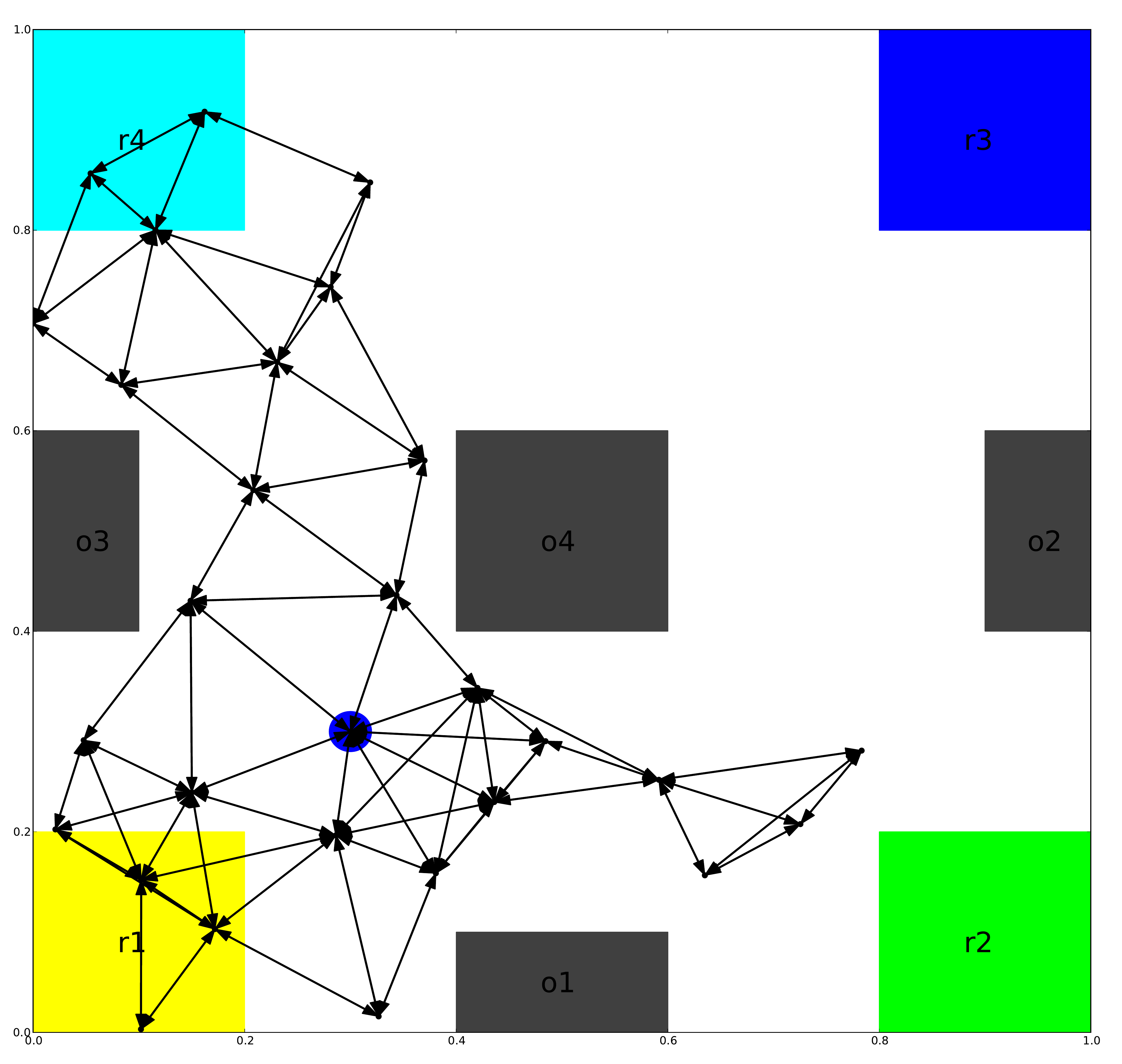}
    \label{fig:case1-s2}
  }
  \subfigure{
    \includegraphics[width=0.2\textwidth]{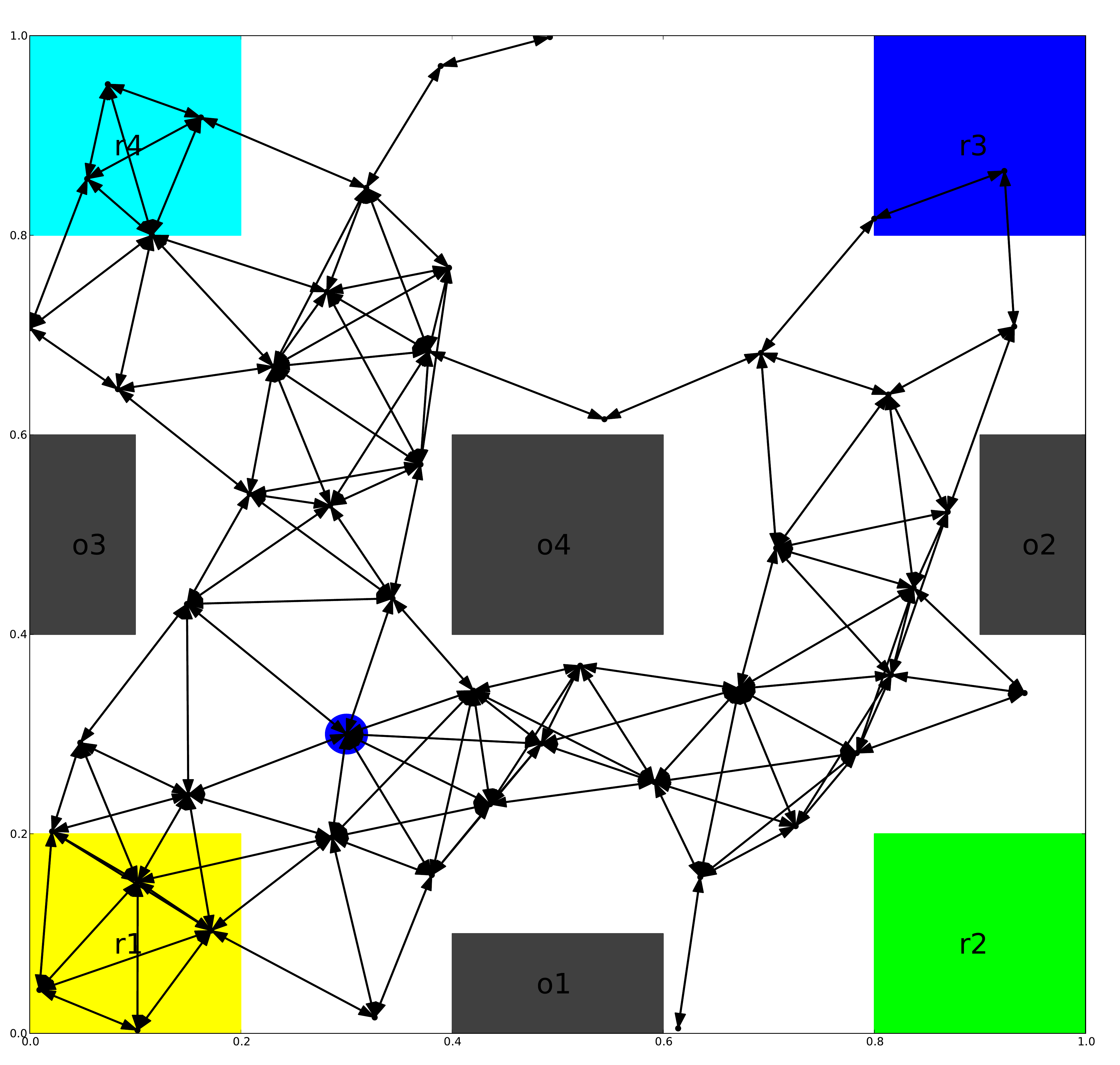}
    \label{fig:case1-s3}
  }
	\caption{Transition systems obtained at earlier iterations corresponding to the
	solution shown in Figure~\ref{fig:case1} (to be read from left to right and top
	to bottom). The black dots and arrows represent the state and transitions of
	$\TS$, respectively.}
	\label{fig:case1-2d}
\end{figure}

A solution to this problem is shown in Figures~\ref{fig:case1}
and~\ref{fig:case1-2d}. We ran the overall algorithm 20 times and obtained an
average execution time of 6.954 sec, out of which the average of the incremental
search algorithm was 6.438 sec. The resulting transition system had a mean size
of 51 states and 277 transitions, while the corresponding product automaton had
a mean size of 643 states  and 7414 transitions. The \buchi automaton
corresponding to $\phi_1$ had 20 states and 155 transitions.

{\bf Case Study 2:} Consider a 10-dimensional unit hypercube configuration
space. The specification is to visit regions $r1$, $r2$, $r3$ infinitely many
times, while avoiding region $o1$. The LTL formula corresponding to this
specification is

\begin{equation}
\phi_2 = \Always( \Event r1 \andltl ( \Event r2 \andltl ( \Event r3 ) ) \andltl \notltl o1 ).
\end{equation}

The corresponding \buchi automaton has 9 states and 43 transitions. Regions
{\cut$r1=[0;0.4] \times [0;0.75]^9$, $r2=[0.6;1]\times[0.25; 1]^9$,
$r3=[0.6;1]\times[0;0.2]\times([0.2;1]\times[0;0.8])^4$ and
$o1=[0.41;59]\times[0.3;0.9]\times[0.12;0.88]^8$} are hypercubes and their
volumes are 0.03, 0.03, 0.013 and 0.012, respectively. $r1$, $r2$, $r3$ are
positioned in the corners of the configuration space, while $o1$ is positioned
in the center. In this case, the algorithm took 16.75 sec on average (20
experiments), while just the incremental search procedure for a satisfying run
took 14.471 sec. The transition system had a mean size of 69 states and 1578
transitions, while the product automaton had a mean size of 439 states and 21300
transitions.

{\bf Case Study 3:} We also considered a 20-dimensional unit hypercube
configuration space. Two hypercube regions $r1$ and $r2$ were defined and the
robot was required to visit both of them infinitely many times ($\phi_3 =
\Always (\Event (r1 \andltl \Event r2))$). The overall algorithm took 7.45
minutes, while the transition system grew to 414 states and 75584 transitions.
The corresponding product automaton had a size of 1145 states and 425544
transitions. {\cut This example illustrates the fact that the bound on the
number of neighbors of a state in the transition system grows at least exponentially in
the dimension of the configuration space.}

{\cut
The performed tests suggest some possible practical improvements of the
execution time of the proposed algorithms. One idea is to postpone the
incremental maintenance of SCCs until there is at least one final state in
$\PA$. We are interested in SCCs only for final states anyway, therefore there
is no benefit to compute them before final states are found. A linear-time SCC
algorithm can be used to initialize the corresponding incremental SCC structure
when final states are found. Another important improvement may be obtained by
processing transitions of $\PA$ in batches. At each iteration, multiple
transitions are added simultaneously, thus a batch version of the incremental
SCC may greatly reduce the total execution time. Note that these heuristics will
not improve the asymptotic bound of the algorithm. Also, the size of the \buchi
automaton has a significant impact on execution time of the procedure, even
though it is fixed and does not contribute to the overall asymptotic bound.
}
{\cut
\section{FUTURE WORK}

As already suggested, future work will include integrating the presented
algorithm with a local on-line sensing and planning procedure. The overall
framework will ensure correctness with respect to both global and local temporal
logic specifications. We will also incorporate realistic robot dynamics and
environment topologies and we will perform experimental validations with air and
ground vehicles in our lab.
}
\bibliography{references}
\bibliographystyle{plain}

\end{document}

%% file: def.tex
\usepackage{amsmath}
\usepackage{amssymb}
\usepackage{amsfonts}
\usepackage{subfigure}
\usepackage{graphicx}
\usepackage{color}
\usepackage{lineno}

\usepackage[linesnumbered,vlined,ruled]{algorithm2e}

\newtheorem{theorem}{Theorem}[section]

\newtheorem{definition}[theorem]{Definition}

\newtheorem{remark}[theorem]{Remark}
\newtheorem{remarks}[theorem]{Remarks}

\newtheorem{problem}[theorem]{Problem}

\newcommand{\todo}[1]{}

\newcommand{\CA}[1]{\mathcal{#1}}
\newcommand{\BF}[1]{\mathbf{#1}}
\newcommand{\BB}[1]{\mathbb{#1}}

\newcommand{\notltl}{\neg}
\newcommand{\andltl}{\wedge}
\newcommand{\orltl}{\vee}
\newcommand{\Next}{\BF{X}}
\newcommand{\Always}{\BF{G}}
\newcommand{\Event}{\BF{F}}
\newcommand{\Until}{\CA{U}}

\newcommand{\buchi}{B\"uchi\ }

\newcommand{\PA}{\mathcal{P}}
\newcommand{\BA}{\mathcal{B}}

\newcommand{\TS}{\mathcal{T}}

\newcommand{\la}{\leftarrow}
\newcommand{\ra}{\rightarrow}
\newcommand{\ras}[1]{\stackrel{#1}{\rightarrow}}
\newcommand{\asgn}{\la}

\newcommand{\normeucl}[1]{\left\| {#1} \right\|_{2}}

\newcommand{\card}[1]{\left| {#1} \right|}
\newcommand{\spow}[1]{2^{#1}}

